\documentclass{article}
\usepackage{spconf,amsmath,graphicx,hyperref}

\usepackage{array}
\usepackage{booktabs}
\usepackage{xcolor}
\usepackage{enumitem}
\usepackage{subcaption} 
\DeclareMathOperator{\argmax}{argmax}

\newcommand{\nhat}[1]{#1} 

\title{RAVE: Rate-Adaptive Visual Encoding for 3D Gaussian Splatting}

  \name{Hoang-Nhat Tran$^{*\dagger}$  Francesco Di Sario$^{*\ddagger}$  Gabriele Spadaro$^{\ddagger\star}$  Giuseppe Valenzise$^\dagger$   Enzo Tartaglione$^\star$}
  \address{$^\dagger$Université Paris-Saclay, CNRS, CentraleSupélec, L2S 91190 Gif-sur-Yvette, France\\
$^\ddagger$University of Turin, Italy\\
$^\star$LTCI, Télécom Paris, Institut Polytechnique de Paris, France\thanks{$^{*}$Co-first authorship\\~\\This article has been accepted for publication at the 
2026 IEEE International Conference on Acoustics, Speech, and Signal Processing (ICASSP).}}
\begin{document}
%
\maketitle
\begin{abstract}
Recent advances in neural scene representations have transformed immersive multimedia, with 3D Gaussian Splatting (3DGS) enabling real-time photorealistic rendering. Despite its efficiency, 3DGS suffers from large memory requirements and costly training procedures, motivating efforts toward compression. Existing approaches, however, operate at fixed rates, limiting adaptability to varying bandwidth and device constraints. In this work, we propose a flexible compression scheme for 3DGS that supports interpolation at any rate between predefined bounds. Our method is computationally lightweight, requires no retraining for any rate, and preserves rendering quality across a broad range of operating points. Experiments demonstrate that the approach achieves efficient, high-quality compression while offering dynamic rate control, making it suitable for practical deployment in immersive applications. The code is available at \url{https://github.com/inspiros/RAVE}.
\end{abstract}
\begin{keywords}
3D Gaussian Splatting, compression, dynamic rate
\end{keywords}
\section{Introduction}
\label{sec:intro}

\begin{figure}[t]

  \centerline{\includegraphics[width=0.8\linewidth]{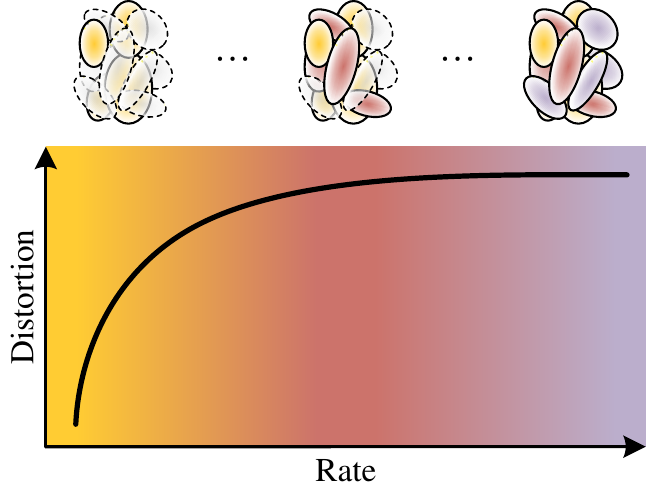}}
%
%
\caption{We present \textbf{RAVE}, the first method enabling \emph{rate-adaptive visual encoding} for 3DGS approaches. Unlike existing methods that require multiple separate training runs, RAVE produces a \emph{continuous} rate–distortion curve in a single, efficient end-to-end training. This allows seamless adaptation of the bitrate to different constraints without retraining, offering both state-of-the-art quality and practical deployment.}

\label{fig:teaser}
\end{figure}

\begin{figure*}[t]
    \centering
    \includegraphics[width=\linewidth]{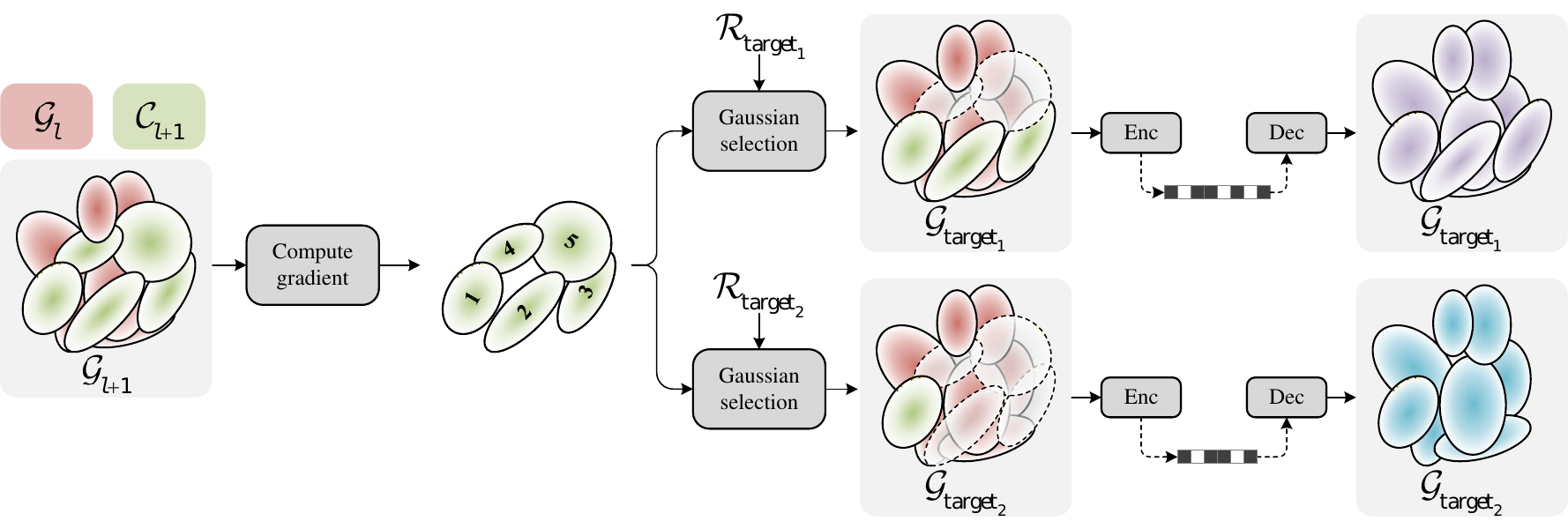}
    \caption{Overview of our method. We compute per-anchor gradient scores once and rank Gaussians by importance. For any target rate $\mathcal{R}_{\text{target}}$, the top-ranked Gaussians are selected to form $\mathcal{G}_{\text{target}}$, then compressed and decoded to reconstruct the scene. This enables multiple operating points and a continuous rate–distortion curve from a single trained model.}
    \label{fig:pipeline}
\end{figure*}

The demand for immersive multimedia experiences has grown rapidly, driven by applications in augmented or virtual reality and interactive entertainment, to mention a few~\cite{hu2025tgavatar}. Realistic scene representation and efficient rendering are now more central than ever to enabling such experiences, where high visual fidelity or realism must be delivered under computational and bandwidth constraints. As a result, research into novel 3D scene representations has recently accelerated, aiming to strike a balance between realism, efficiency, and scalability~\cite{ali2025compression}.
\\
A pivotal advance in this area was introduced by neural radiance fields (NeRFs)~\cite{mildenhall2021nerf},
which demonstrated that neural implicit functions could effectively synthesize photorealistic views from sparse imagery. 
Despite their success, NeRFs suffer from substantial computational overhead in both training and rendering.
More recently, 3D Gaussian Splatting (3DGS)~\cite{kerbl20233d} has emerged as a powerful alternative, relying on explicit point-based representations to enable real-time rendering. This shift has further boosted the field~\cite{wu2024recent,zou2025gaussianenhancer}, offering a practical pathway toward interactive immersive media with an end-to-end learning fashion.
\\
Nevertheless, 3DGS also faces inherent challenges.
While its explicit representation allows for fast rendering~\cite{hanson2025speedy}, the memory footprint for rendering can grow significantly~\cite{ali2025elmgs}. 
Beyond efficiency, scaling the training process remains computationally demanding, 
often requiring significant optimization steps to achieve competitive quality~\cite{zhao2024scaling}.
These issues hinder widespread adoption in resource-constrained environments, where lightweight yet flexible solutions are crucial.
\\
To overcome these limitations, several compression techniques have been proposed to reduce both storage and transmission costs.
In this setting, a common approach is to prune Gaussians, thereby reducing the overall representation size~\cite{ali2025compression}.
However, most existing methods operate at fixed compression rates~\cite{chen2024hac, wang2024contextgs, liu2024compgs, 3DGSzip2024}, 
which limits their adaptability to different application requirements, such as variable bandwidth and device capabilities. 
This rigidity limits the practical deployment of 3DGS across various real-world scenarios, where dynamic trade-offs between quality, storage, and computation are often required in real-time.
\\
In this paper, we tackle this issue by proposing RAVE, a rate-adaptive model-agnostic method that enables flexible compression of 3DGS representations.
Unlike previous approaches, RAVE supports interpolation at any rate within a specified range, yielding fine-grained control over the rate-distortion trade-off through predefined contexts (Fig.~\ref{fig:teaser}). 
With marginal extra training costs (in terms of learning iterations), our method is the \emph{first to allow dynamic rate adaptation without retraining}.
Notably, achieving such flexibility is not straightforward, as naïve solutions tend to either degrade quality or impose significant computational overhead. 

\begin{figure*}[t]
    \renewcommand{\arraystretch}{.5}
    \centering
    \small
    \begin{tabular}{@{}c@{}c@{}c@{}}
    \qquad\quad\textbf{Mip-NeRF 360} & \qquad\quad\textbf{Tanks\&Temples} & \qquad\textbf{Deep Blending} \\
    \begin{subfigure}{0.32\textwidth}
        \includegraphics[width=\linewidth]{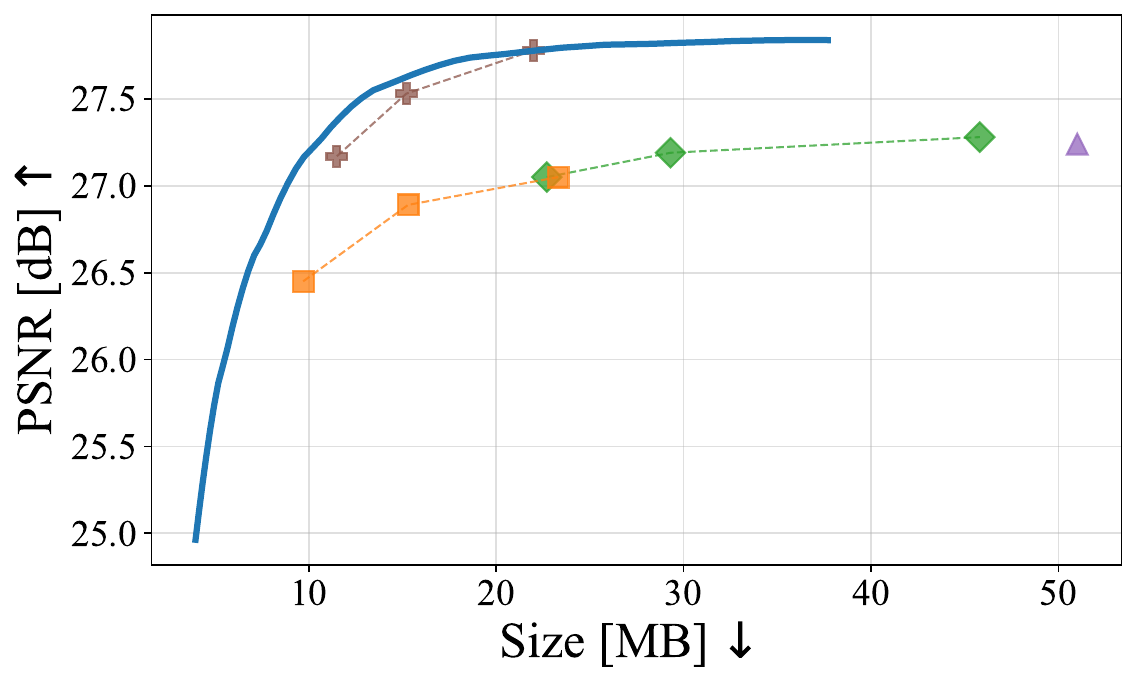}
    \end{subfigure} & %
    \begin{subfigure}{0.32\textwidth}
        \includegraphics[width=\linewidth]{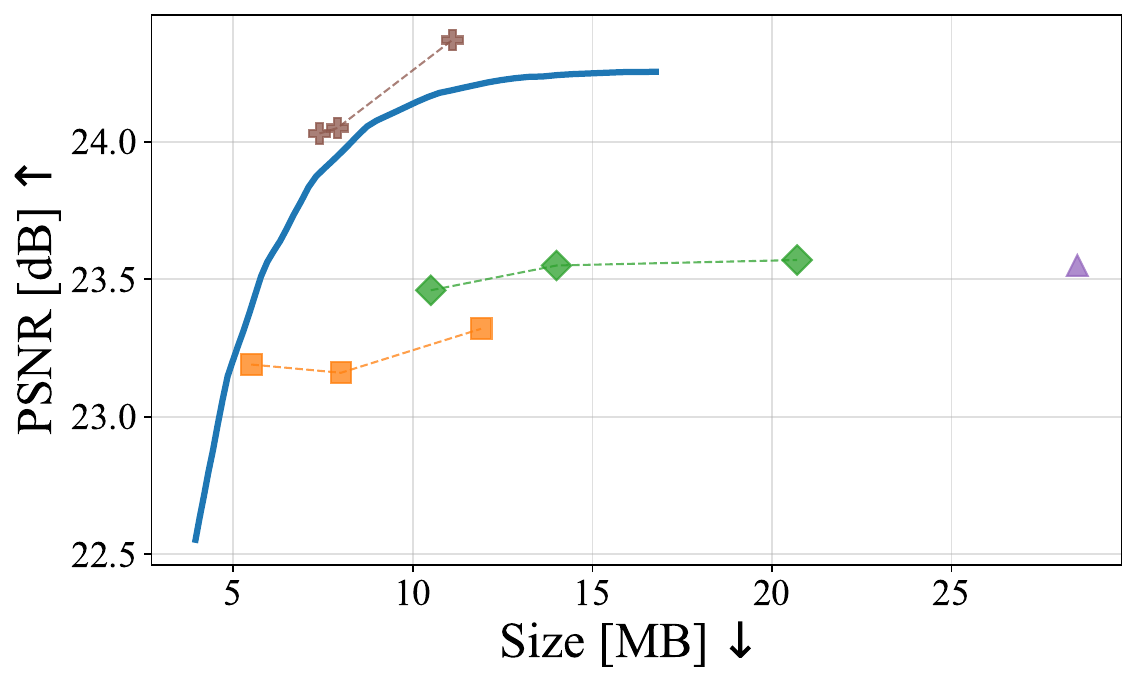}
    \end{subfigure} & %
    \begin{subfigure}{0.32\textwidth}
        \includegraphics[width=\linewidth]{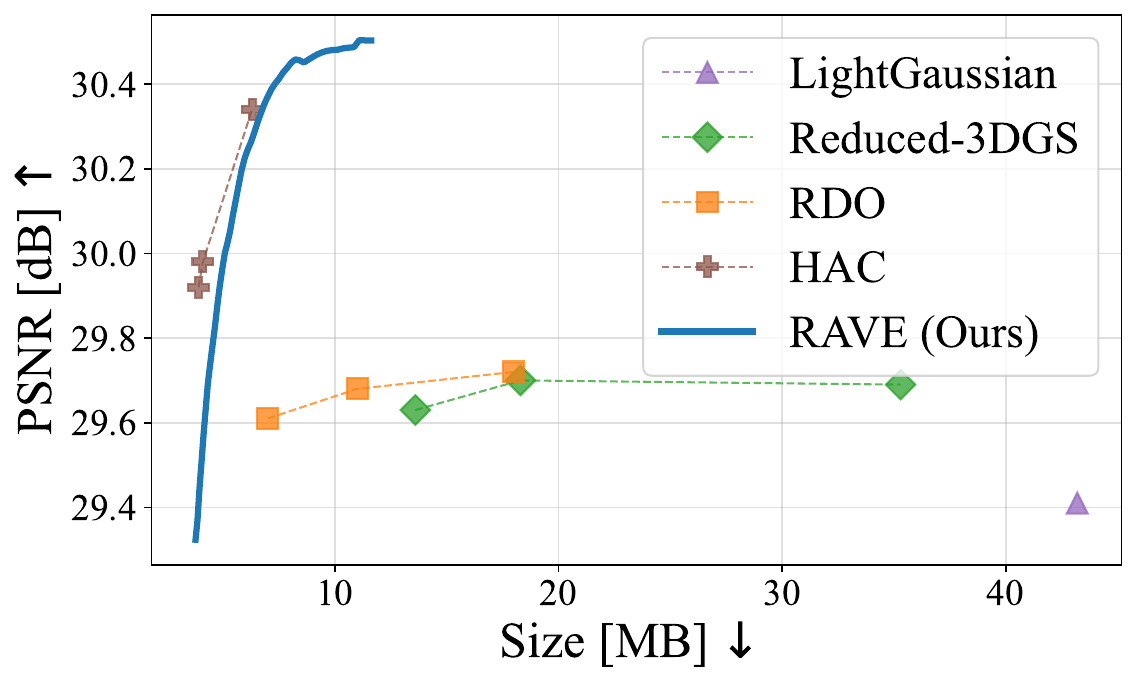}
    \end{subfigure} \\
    \begin{subfigure}{0.32\textwidth}
        \includegraphics[width=\linewidth]{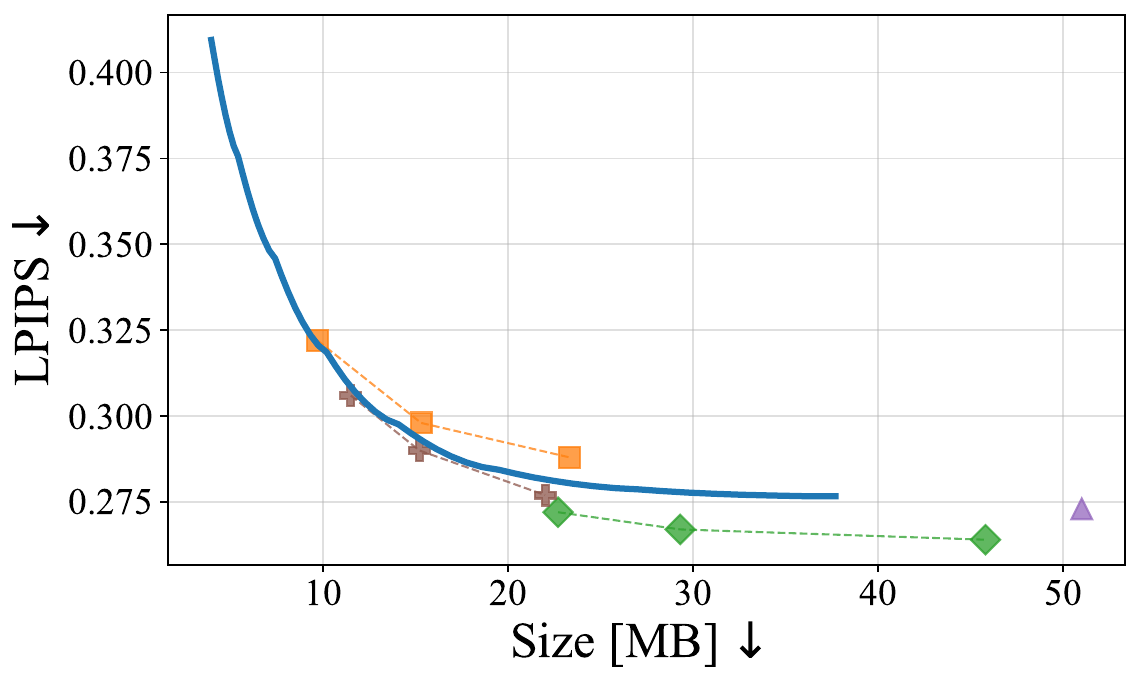}
    \end{subfigure} & %
    \begin{subfigure}{0.32\textwidth}
        \includegraphics[width=\linewidth]{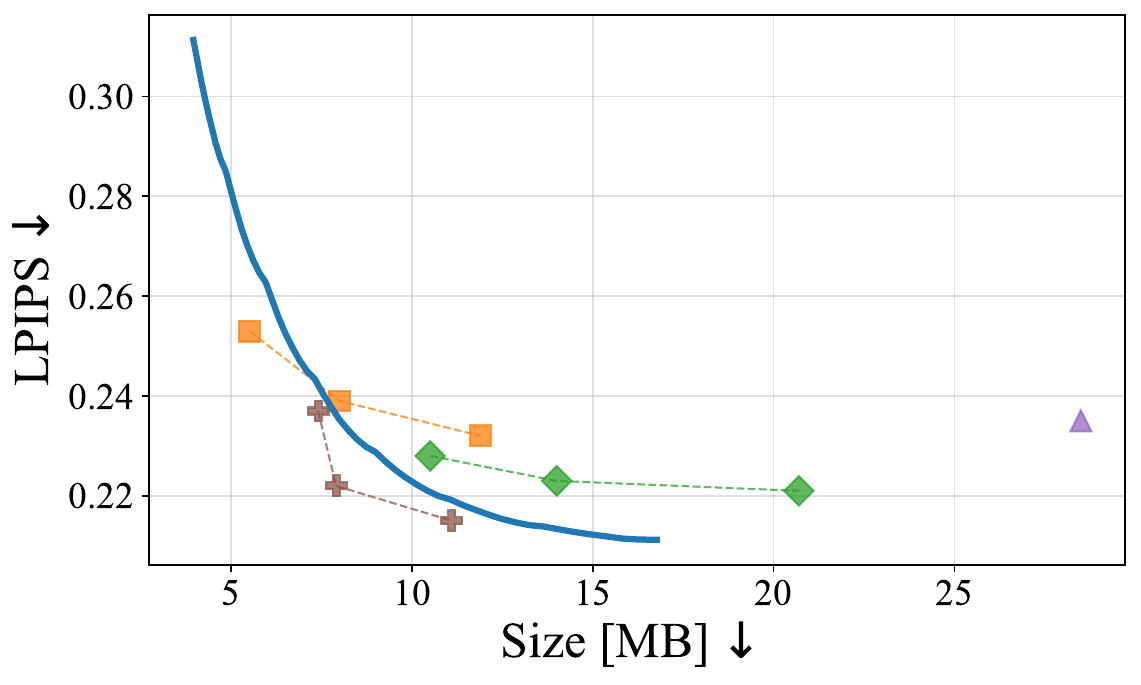}
    \end{subfigure} & %
    \begin{subfigure}{0.32\textwidth}
        \includegraphics[width=\linewidth]{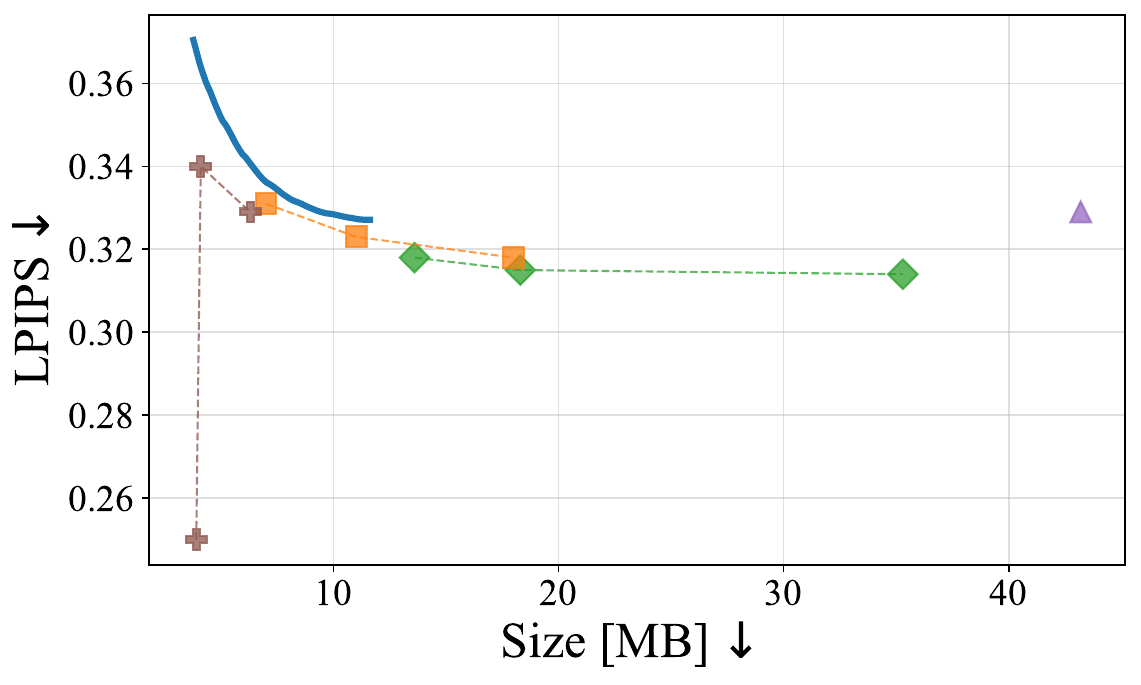}
    \end{subfigure}
    \end{tabular}
    \vspace{-0.5em}
    \caption{Comparison of RAVE against state-of-the-art compression methods \nhat{in PSNR-Size (top) and LPIPS-Size (bottom)}. Our method is the only one that allows for a continuous rate without requiring any retraining. The results are competitive, achieving state-of-the-art performance on \emph{Mip-NeRF 360}. All other methods require a separate training run for each point on the curve.}
    \label{fig:main}
\end{figure*}

\section{Rate-adaptive Gaussian Splatting}
\label{sec:format}

\subsection{Preliminaries}
3D Gaussian Splatting~\cite{kerbl20233d} has emerged as a state-of-the-art method for real-time radiance field rendering, combining high image quality with exceptional rendering speed.
In this approach, a scene is represented by a set of volumetric Gaussian primitives, each defined by a position $x$, a covariance matrix $\Sigma$ (decomposed into scale and rotation components), an opacity $o$, and a set of Spherical Harmonics (SH) coefficients that encode view-dependent color or appearance. 
During the rendering stage, 3D Gaussians are projected into the 2D image plane and alpha-blended. Blending weights for a given projected point are computed as:
\begin{equation}
\alpha(x) = o \cdot G(x),
\text{~~~~with~~~~} 
G(x) = \exp\!\left(-\frac{1}{2} x^{T} \Sigma^{-1} x \right)
\end{equation}
representing the Gaussian contribution at position $x$.
The SH coefficients are evaluated to produce color values specific to the current view, and the combination of these colors through alpha compositing reconstructs the final appearance of the scene.
The initial set of Gaussian primitives is derived from a sparse Structure-from-Motion (SfM) point cloud, providing a coarse scene initialization.
These parameters are then optimized through gradient descent, while simultaneously performing adaptive refinement operations that split, merge, or adjust the primitives to better match the underlying geometry.
The optimization jointly minimizes pixel-wise reconstruction error and structural dissimilarity.

\subsection{Training for variable rate transmission}
Taking inspiration from the literature on \nhat{classical scalable video coding~\cite{li2001mpeg4, schwarz2007svc}} and Level of Details for 3DGS~\cite{di2025gode}, we can introduce a variable-rate approach by defining anchor points at $L$ fixed rate-distortion levels. Starting from a pre-trained model, a hierarchy of gradient-based masks can be generated to progressively select subsets of Gaussians. Formally, the hierarchy is defined as \nhat{$\mathcal{G}_l = \bigcup_{i=1}^{l} \mathcal{C}_i$, with $\mathcal{G}_l$ grouping Gaussians from the lowest rate anchor to $l$, and $\mathcal{C}_l$ representing our \emph{context} comprising Gaussians to be introduced to move from anchor $l-1$ to $l$}. The Gaussians hierarchy is established by defining a rate delta $\Delta \mathcal{R}$  that translates to a fraction of Gaussians to be added/removed to move from anchor $l-1$ to $l$, and their ranking is defined based on the gradient calculated on the training set.
\\
After building the hierarchy, we can perform quantization-aware fine-tuning using stochastic mask sampling: this means that, with uniform probability, we choose to optimize either of the $L$ anchor points.
\\
Once fine-tuned, we can easily interpolate \nhat{between any anchor pairs $l$ and $l + 1$ by re-estimating the gradient on the subset $\mathcal{C}_{l + 1}$}: leveraging the principle of locality, we know that the denser the anchors for the same bitrate regime, the more accurate the gradient-based ranking.
Therefore, given $\mathcal{R}(\mathcal{G}_l)$ the rate for $\mathcal{G}_l$, the number of Gaussians for a specific bitrate $\mathcal{R}_{\text{target}}$, for which $l=\argmax\{\mathcal{R}(\mathcal{G}_l)\leq \mathcal{R}_{\text{target}}\}$, will be
\begin{equation}
    \left|\mathcal{G}_{\text{target}}\right| = \left|\mathcal{G}_{l}\right|+ \frac{\mathcal{R}_{\text{target}} - \mathcal{R}(\mathcal{G}_l)}{\mathcal{R}(\mathcal{G}_{l+1})-\mathcal{R}(\mathcal{G}_l)}(|\mathcal{G}_{l+1}| - |\mathcal{G}_l|) .
\end{equation}
Among the budget $ |\Delta \mathcal{G}| = \left|\mathcal{G}_{\text{target}}\right| - \left|\mathcal{G}_{l}\right|$, we select the Gaussians having the highest gradient from the context $\mathcal{C}_{l+1}$:
\begin{equation}
    \Delta \mathcal{G} = \bigcup_{i=1}^{ |\Delta \mathcal{G}|}G_i \left|G_i\in \mathcal{C}_{l+1}, \left \lVert\frac{\partial \mathcal{L}}{\partial \theta_i} \right\rVert_2 \nhat{\geq} \left\lVert\frac{\partial \mathcal{L}}{\partial \theta_j} \right\rVert_2 \forall i < j\right . ,
\end{equation}
where $\theta$ represents the parameters of the Gaussian, and $\mathcal{L}$ is the rendering loss function. The gradients are calculated only once on the training set.

\subsection{Method overview}
In Fig.~\ref{fig:pipeline}, we present an overview of our method.
Our objective is to generate high-quality scene reconstructions at an arbitrary target rate $\mathcal{R}_{\text{target}}$, determined by constraints such as network bandwidth, storage budget, or hardware capabilities, without requiring additional training or model fine-tuning.
We start by considering a set of Gaussians $\mathcal{G}_{l}$ produced at a given anchor level, together with its local context $\mathcal{C}_{l+1}$.
Each anchor acts as a local reference, restricting the optimization to a well-defined subset of Gaussians and allowing for efficient computation.
For each Gaussian, we compute an importance score by accumulating its gradient over a single forward pass through the training set. This operation is performed only once per anchor level and provides a stable ranking of Gaussians based on their contribution to reconstruction quality.
Given a target rate $\mathcal{R}_{\text{target}}$, we identify the anchor level that contains the desired number of Gaussians and select the top-ranked ones until the rate budget is satisfied, producing a subset $\mathcal{G}_{\text{target}}$.
Crucially, since the ranking is precomputed, this selection process can be repeated for any number of target rates $\mathcal{R}_{\text{target}_1}, \mathcal{R}_{\text{target}_2}, \dots$ without recomputing gradients, enabling the generation of multiple rate–quality operating points from a single trained model.
Finally, the parameters of the selected Gaussians are compressed using entropy coding (LZMA in our implementation, chosen for its strong compression ratio and practical speed), transmitted, and decoded at the receiver side to reconstruct the scene.
Because our method is codec-agnostic, different compression backends could be substituted with no modification to the overall pipeline.
This design allows us to generate an entire continuous rate–distortion curve from a single training run. 


\section{EXPERIMENTS}
\label{sec:pagestyle}


\begin{figure}[t]
    \centering
    \includegraphics[width=.88\linewidth]{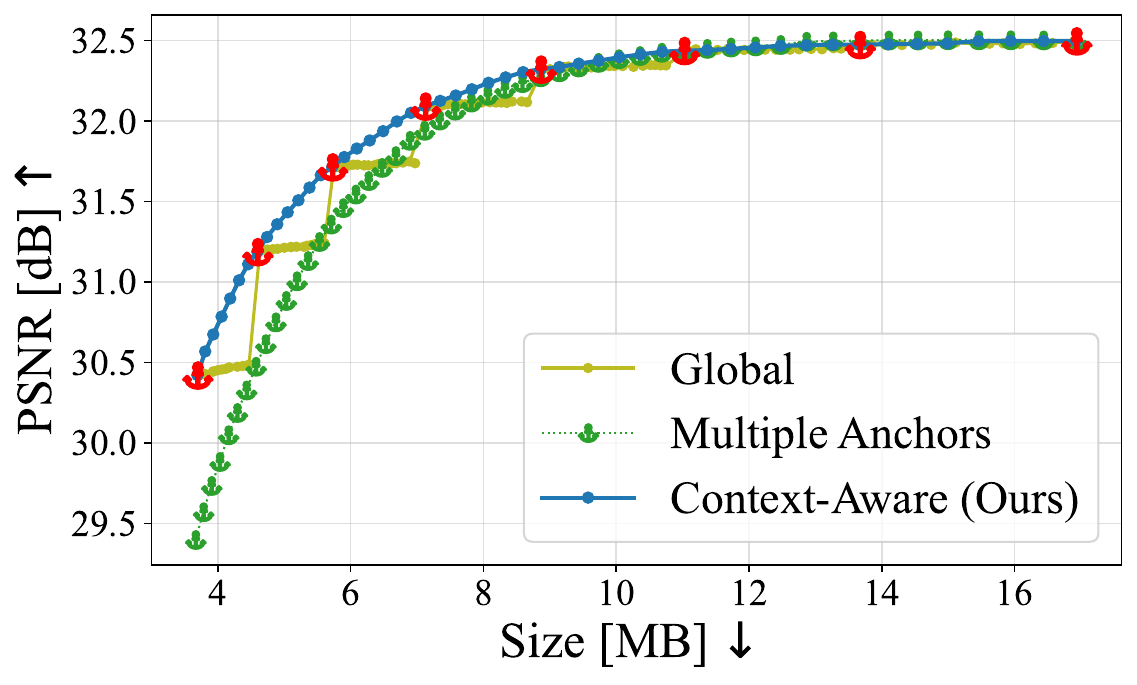}
    \caption{Comparison with a global variant of our method, where the gradient is computed globally before each pruning stage, and with a naïve multi-anchors strategy (50 levels). Our method highlights the importance of context, while at the same time being superior to the multi-anchor strategy.}
    \label{fig:ablation}
\end{figure}

\subsection{Setup}
We evaluated our method on the three major benchmarks for 3D Gaussian Splatting: \emph{Mip-NeRF 360}~\cite{barron2022mip}, \emph{Tanks\&\allowbreak Temples}~\cite{knapitsch2017tanks}, and \emph{Deep Blending}~\cite{DeepBlending2018}. These datasets cover diverse real-world scenarios and provide challenging conditions for assessing reconstruction quality and generalization. For the 3DGS model, we adopt Scaffold-GS~\cite{scaffoldgs} as our backbone architecture.
\\
To perform Gaussian selection, we compute the gradient with respect to \nhat{Gaussian feature parameters as in~\cite{di2025gode}, considering the rendering loss $\mathcal{L}$ consisting of a pixel-wise $\mathcal{L}_1$ error and} an SSIM term $\mathcal{L}_\mathrm{SSIM}$ as in standard 3DGS-based methods.
\begin{equation}
    \mathcal{L} = (1-\lambda) \mathcal{L}_1 + \lambda \mathcal{L}_\mathrm{SSIM},
\end{equation}
where $\lambda$ is a unit interval ratio hyperparameter set to $0.2$ in our experiments.

\subsection{Main results}

As shown in Fig.~\ref{fig:main}, our method achieves competitive performance in both PSNR and LPIPS compared to the most recent 3DGS compression techniques.
Remarkably, RAVE is the only approach that supports continuous variable rate compression, whereas all other single-rate methods require separate, ad hoc training runs to obtain each point on the rate–distortion curve.
\nhat{Training multiple models for different rates} is very expensive and, hence, not feasible in practice, where we often need to adapt the rate to specific constraints.
More specifically, our method reaches state-of-the-art performance on \emph{Mip-NeRF 360}, slightly outperforming HAC, while still being competitive on the other datasets. 
\nhat{
It is particularly noteworthy that as we reduce the size of the lowest level $\mathcal{G}_1$, our method is able to achieve arbitrarily low bitrate operating points.
In extreme low-bitrate regimes, removing Gaussians significantly degrades the reconstruction quality, leading to steeper regions in the left-most parts of the RD curves.
}


\subsection{Variable rate transmission is not easy}

\textbf{Global vs Context-Aware.} 
To better understand the role of the interpolation strategy, we consider two distinct approaches.
In the first case, the ordering of the newly introduced Gaussians (those present in level $l+1$ but not in level $l$) is derived from gradients computed with respect to the \emph{global} scene, \textit{i.e.}, the reconstruction in which all Gaussians up to the highest-rate anchor are included.
With this global context, however, the ranking is dominated by structures that become relevant only at higher rates, resulting in poorly aligned intermediate reconstructions.
Fig.~\ref{fig:ablation} shows the results obtained in the \emph{room} scene.
The RD curve obtained in this way \nhat{(olive curve)} exhibits irregularities, thus preventing a smooth transition between the anchors at different levels.
In our approach, instead, we restrict the ranking and selection to a local context $\mathcal{C}_{l+1}$.
This subtle change is crucial, as it ensures that the ranking reflects the contribution of the Gaussians added at that specific target rate, resulting in a smooth RD curve with more coherently evolving intermediate models.


\noindent
\textbf{Multiple anchors.}
We also compare our approach against a naïve strategy (green curve in Fig.~\ref{fig:ablation}), which attempts to attain a continuous RD curve by defining an arbitrary number of anchors.
In principle, with this simple strategy, we can densely populate the rate axis.
However, this comes at a significant computational cost.
Indeed, it is necessary to compute a new ordering of the Gaussians based on gradient evaluation for each new anchor.
As the number of levels increases, this calculation grows linearly more expensive, thereby making this strategy impractical for real-world applications.
\\
Moreover, as shown in Fig.~\ref{fig:ablation}, increasing the number of levels (\textit{e.g.}, up to $50$ in our experiment) deteriorates the RD behavior. 
This highlights a structural limitation of such an approach: while being able to approximate continuity by brute force, it becomes both inefficient and unstable as the number of anchors grows.
In contrast, our approach achieves a truly continuous RD curve while being faster, more lightweight, and consistently yielding higher quality reconstructions.

\section{CONCLUSION}
\label{sec:conclusion}

We presented a flexible compression framework for 3D Gaussian Splatting that allows continuous interpolation between compression rates while maintaining efficient rendering and competitive visual quality. Unlike existing approaches constrained to fixed-rate operation, our method adapts seamlessly to diverse computational and bandwidth conditions, enabling practical deployment in immersive multimedia scenarios.
The results show that high-quality rendering can be preserved across a wide range of operating points at minimal computational cost, demonstrating that adaptive compression of 3DGS is both feasible and effective. This opens the door to more versatile scene representations that can dynamically respond to application requirements, paving the way toward scalable, interactive, and resource-aware immersive experiences.


\section{Acknowledgements}
This work was supported by the French National Research Agency (ANR) in the framework of the IA Cluster project “Hi! PARIS Cluster 2030” under Grant ANR-23-IACL-005, and by the Hi! PARIS Center on Data Analytics and Artificial Intelligence.


\bibliographystyle{IEEEbib}
\bibliography{main}

@article{mildenhall2021nerf,
  title={Nerf: Representing scenes as neural radiance fields for view synthesis},
  author={Mildenhall, Ben and Srinivasan, Pratul P and Tancik, Matthew and Barron, Jonathan T and Ramamoorthi, Ravi and Ng, Ren},
  journal={Communications of the ACM},
  volume={65},
  number={1},
  pages={99--106},
  year={2021},
  publisher={ACM New York, NY, USA}
}

@article{kerbl20233d,
  title={3D Gaussian splatting for real-time radiance field rendering.},
  author={Kerbl, Bernhard and Kopanas, Georgios and Leimk{\"u}hler, Thomas and Drettakis, George},
  journal={ACM Trans. Graph.},
  volume={42},
  number={4},
  pages={139--1},
  year={2023}
}

@article{wu2024recent,
  title={Recent advances in 3d gaussian splatting},
  author={Wu, Tong and Yuan, Yu-Jie and Zhang, Ling-Xiao and Yang, Jie and Cao, Yan-Pei and Yan, Ling-Qi and Gao, Lin},
  journal={Computational Visual Media},
  volume={10},
  number={4},
  pages={613--642},
  year={2024},
  publisher={TUP}
}

@inproceedings{zou2025gaussianenhancer,
  title={GaussianEnhancer: A General Rendering Enhancer for Gaussian Splatting},
  author={Zou, Chen and Ma, Qingsen and Wang, Jia and Lu, Ming and Zhang, Shanghang and He, Zhaofeng},
  booktitle={ICASSP 2025-2025 IEEE International Conference on Acoustics, Speech and Signal Processing (ICASSP)},
  pages={1--5},
  year={2025},
  organization={IEEE}
}

@inproceedings{hanson2025speedy,
  title={Speedy-splat: Fast 3d gaussian splatting with sparse pixels and sparse primitives},
  author={Hanson, Alex and Tu, Allen and Lin, Geng and Singla, Vasu and Zwicker, Matthias and Goldstein, Tom},
  booktitle={Proceedings of the Computer Vision and Pattern Recognition Conference},
  pages={21537--21546},
  year={2025}
}

@inproceedings{ali2025elmgs,
  title={Elmgs: Enhancing memory and computation scalability through compression for 3d gaussian splatting},
  author={Ali, Muhammad Salman and Bae, Sung-Ho and Tartaglione, Enzo},
  booktitle={2025 IEEE/CVF Winter Conference on Applications of Computer Vision (WACV)},
  pages={2591--2600},
  year={2025},
  organization={IEEE}
}

@inproceedings{zhao2024scaling,
  title={On scaling up 3d gaussian splatting training},
  author={Zhao, Hexu and Weng, Haoyang and Lu, Daohan and Li, Ang and Li, Jinyang and Panda, Aurojit and Xie, Saining},
  booktitle={European Conference on Computer Vision},
  pages={14--36},
  year={2024},
  organization={Springer}
}

@article{ali2025compression,
  title={Compression in 3d gaussian splatting: A survey of methods, trends, and future directions},
  author={Ali, Muhammad Salman and Zhang, Chaoning and Cagnazzo, Marco and Valenzise, Giuseppe and Tartaglione, Enzo and Bae, Sung-Ho},
  journal={arXiv preprint arXiv:2502.19457},
  year={2025}
}

@article{hu2025tgavatar,
  title={TGAvatar: Reconstructing 3D Gaussian Avatars with Transformer-based Tri-plane},
  author={Hu, Ruigang and Wang, Xuekuan and Yan, Yichao and Zhao, Cairong},
  journal={IEEE Transactions on Circuits and Systems for Video Technology},
  year={2025},
  publisher={IEEE}
}

@String(CVPR= {IEEE Conf. Comput. Vis. Pattern Recog.})

@String(TOG= {ACM Trans. Graph.})

@String(ICASSP=	{ICASSP})

@String(CVPR  = {CVPR})

@String(TOG   = {ACM TOG})

@inproceedings{barron2022mip,
  title={Mip-nerf 360: Unbounded anti-aliased neural radiance fields},
  author={Barron, Jonathan T and Mildenhall, Ben and Verbin, Dor and Srinivasan, Pratul P and Hedman, Peter},
  booktitle={Proceedings of the IEEE/CVF Conference on Computer Vision and Pattern Recognition},
  pages={5470--5479},
  year={2022}
}

@inproceedings{chen2024hac,
  title={HAC: Hash-grid Assisted Context for 3D Gaussian Splatting Compression},
  author={Chen, Yihang and Wu, Qianyi and Lin, Weiyao and Harandi, Mehrtash and Cai, Jianfei},
  booktitle={European Conference on Computer Vision},
  year={2024}
}

@misc{3DGSzip2024,
    title={3DGS.zip: A survey on 3D Gaussian Splatting Compression Methods}, 
    author={Milena T. Bagdasarian and Paul Knoll and Yi-Hsin Li and Florian Barthel and Anna Hilsmann and 
            Peter Eisert and Wieland Morgenstern},
    year={2024},
    eprint={2407.09510},
    archivePrefix={arXiv},
    primaryClass={cs.CV},
    url={https://arxiv.org/abs/2407.09510}, 
}

@InProceedings{scaffoldgs,
    author    = {Lu, Tao and Yu, Mulin and Xu, Linning and Xiangli, Yuanbo and Wang, Limin and Lin, Dahua and Dai, Bo},
    title     = {Scaffold-GS: Structured 3D Gaussians for View-Adaptive Rendering},
    booktitle = {Proceedings of the IEEE/CVF Conference on Computer Vision and Pattern Recognition (CVPR)},
    month     = {June},
    year      = {2024},
    pages     = {20654-20664}
}

@article{knapitsch2017tanks,
  title={Tanks and temples: Benchmarking large-scale scene reconstruction},
  author={Knapitsch, Arno and Park, Jaesik and Zhou, Qian-Yi and Koltun, Vladlen},
  journal={ACM Transactions on Graphics (ToG)},
  volume={36},
  number={4},
  pages={1--13},
  year={2017},
  publisher={ACM New York, NY, USA}
}

@article{DeepBlending2018,
  author = {Hedman, Peter and Philip, Julien and Price, True and Frahm, Jan-Michael and Drettakis, George and Brostow, Gabriel},
  title = {Deep Blending for Free-viewpoint Image-based Rendering},
  booktitle = {ACM Transactions on Graphics (Proc. SIGGRAPH Asia)},
  publisher = {ACM},
  volume    = {37},
  number    = {6},
  pages     = {257:1--257:15},
  year      = {2018}
}

@article{wang2024contextgs,
  title={Contextgs: Compact 3d gaussian splatting with anchor level context model},
  author={Wang, Yufei and Li, Zhihao and Guo, Lanqing and Yang, Wenhan and Kot, Alex and Wen, Bihan},
  journal={Advances in neural information processing systems},
  volume={37},
  pages={51532--51551},
  year={2024}
}

@inproceedings{liu2024compgs,
  title={Compgs: Efficient 3d scene representation via compressed gaussian splatting},
  author={Liu, Xiangrui and Wu, Xinju and Zhang, Pingping and Wang, Shiqi and Li, Zhu and Kwong, Sam},
  booktitle={Proceedings of the 32nd ACM International Conference on Multimedia},
  pages={2936--2944},
  year={2024}
}

@article{di2025gode,
  title={Gode: Gaussians on demand for progressive level of detail and scalable compression},
  author={Di Sario, Francesco and Renzulli, Riccardo and Grangetto, Marco and Sugimoto, Akihiro and Tartaglione, Enzo},
  journal={arXiv preprint arXiv:2501.13558},
  year={2025}
}

@ARTICLE{li2001mpeg4,
  author={Weiping Li},
  journal={IEEE Transactions on Circuits and Systems for Video Technology}, 
  title={Overview of fine granularity scalability in MPEG-4 video standard}, 
  year={2001},
  volume={11},
  number={3},
  pages={301-317},
  doi={10.1109/76.911157}
}

@article{schwarz2007svc,
  author={Schwarz, Heiko and Marpe, Detlev and Wiegand, Thomas},
  journal={IEEE Transactions on Circuits and Systems for Video Technology}, 
  title={Overview of the Scalable Video Coding Extension of the H.264/AVC Standard}, 
  year={2007},
  volume={17},
  number={9},
  pages={1103-1120},
  doi={10.1109/TCSVT.2007.905532}
}

\end{document}